\documentclass[letterpaper]{article}
\usepackage{aaai20}
\usepackage{times}
\usepackage{helvet}
\usepackage{courier}
\usepackage[hyphens]{url}
\usepackage{graphicx}
\usepackage{booktabs}
\usepackage{times}  
\usepackage{url}
\usepackage{multirow,multicol}
\usepackage{amsmath,amssymb,amscd,epsfig,amsfonts,rotating}
\usepackage{graphicx}
\usepackage{multirow}
\usepackage{booktabs}
\usepackage{amsmath,amscd,amsbsy,amssymb,latexsym,url,bm}
\usepackage{epsfig,epstopdf,graphicx,subfigure}

\begin{document}
%
\title{Symmetric Regularization based BERT for Pair-wise Semantic Reasoning
}
\makeatletter
\newcommand{\printfnsymbol}[1]{%
  \textsuperscript{\@fnsymbol{#1}}%
}
\makeatother

\author{ 
Weidi Xu~\textsuperscript{\rm 1}\printfnsymbol{1}, 
Xingyi Cheng~\textsuperscript{\rm 1}\thanks{Equal Contribution.},
Kunlong Chen~\textsuperscript{\rm 1}\printfnsymbol{1}, 
Wei Wang~\textsuperscript{\rm 2}, 
Bin Bi~\textsuperscript{\rm 2}, \\
\Large \textbf{Ming Yan~\textsuperscript{\rm 2}, 
Chen Wu~\textsuperscript{\rm 2}, 
Luo Si~\textsuperscript{\rm 2}, 
Wei Chu~\textsuperscript{\rm 1} \and 
Taifeng Wang~\textsuperscript{\rm 1}}\\
~\textsuperscript{\rm 1} Ant Financial Services Group \\
~\textsuperscript{\rm 2} DAMO, Alibaba Group \\
{\rm \{weidi.xwd,fanyin.cxy,kunlong.ckl,hebian.ww,b.bi,ym119608,wuchen.wc,luo.si,
weichu.cw,taifeng.wang\}@alibaba-inc.com}
  }
\date{}

  %

\maketitle
\begin{abstract}
\begin{quote}
The ability of semantic reasoning over the sentence pair is essential for many natural language understanding tasks, e.g., natural language inference and machine reading comprehension.
A recent significant improvement in these tasks comes from BERT.
As reported, the next sentence prediction (NSP) in BERT, which learns the contextual relationship between two sentences, is of great significance for downstream problems with sentence-pair input.
Despite the effectiveness of NSP, we suggest that NSP still lacks the essential signal to distinguish between entailment and shallow correlation.
To remedy this, we propose to augment the NSP task to a 3-class categorization task, which includes a category for previous sentence prediction (PSP).
The involvement of PSP encourages the model to focus on the informative semantics to determine the sentence order, thereby improves the ability of semantic understanding.
This simple modification yields remarkable improvement against vanilla BERT.
To further incorporate the document-level information, the scope of NSP and PSP is expanded into a broader range, i.e., NSP and PSP also include close but nonsuccessive sentences, the noise of which is mitigated by the label-smoothing technique.
Both qualitative and quantitative experimental results demonstrate the effectiveness of the proposed method.
Our method consistently improves the performance on the NLI and MRC benchmarks, including the challenging HANS dataset~\cite{hans}, suggesting that the document-level task is still promising for the pre-training.
\end{quote}
\end{abstract}

\section{Introduction}
\noindent 
 ~\footnote{We withdrew the paper in ACL2020. The code was release at {https://github.com/ACL2020Anonymous/SymBERT} in Dec. 2019.}
The ability of semantic reasoning is essential for advanced natural language understanding (NLU) systems.
Many NLU tasks that take sentence pairs as input, such as natural language inference (NLI) and machine reading comprehension (MRC), heavily rely on the ability of sophisticated semantic reasoning.
For instance, the NLI task aims to determine whether the hypothesis sentence (e.g., a woman is sleeping) can be inferred from the premise sentence (e.g., a woman is talking on the phone).
This requires the model to read and understand sentence pairs to make the specific semantic inference.

Bidirectional Encoder Representations from Transformer (BERT)~\cite{devlin2018bert} has shown strong ability in semantic reasoning.
It was recently proposed and obtained impressive results on many tasks, ranging from text classification, natural language inference, and machine reading comprehension. 
BERT achieves this by employing two objectives in the pre-training, i.e., the masked language modeling (Masked LM) and the next sentence prediction (NSP). 
Intuitively, the Masked LM task concerns word-level knowledge, and the NSP task captures the global document-level information.
The goal of NSP is to identify whether an input sentence is next to another input sentence.
From the ablation study~\cite{devlin2018bert}, the NSP task is quite useful for the downstream NLI and MRC tasks (e.g., +3.5\% absolute gain on the Question NLI (QNLI)~\cite{glue} task).

Despite its usefulness, we suggest that BERT has not made full use of the document-level knowledge.
The sentences in the negative samples used in NSP are randomly drawn from other documents.
Therefore, to discriminate against these sentences, BERT is prone to aggregating the shallow semantic, e.g., topic, neglecting context clues useful for detailed reasoning.
In other words, the canonical NSP task would encourage the model to recognize the correlation between sentences, rather than obtaining the ability of semantic entailment. 
This setting weakens the BERT model from learning specific semantic for inference.
Another issue that renders NSP less effective is that BERT is order-sensitive.
Performance degradation was observed on typical NLI tasks when the order of two input sentences are reversed during the BERT fine-tuning phase.
It is reasonable as the NSP task can be roughly analogy to the NLI task when the input comes as (premise, hypothesis), considering the causal order among sentences.
However, this identity between NSP and NLI is compromised when the sentences are swapped. 

Based on these considerations, we propose a simple yet effective method, i.e., introducing a \texttt{IsPrev} category to the classification task, which is a symmetric label of \texttt{IsNext} of NSP.
The input of samples with \texttt{IsPrev} is the reverse of those with \texttt{IsNext} label.
The advantages of using this previous sentence prediction (PSP) are three folds. 
(1) Learning the contrast between NSP and PSP forces the model to extract more detailed semantic, thereby the model is more capable of discriminating the correlation and entailment. 
(2) NSP and PSP are symmetric. This symmetric regularization alleviates the influence of the order of the input pair. 
(3) Empirical results indicate that our method is beneficial for all the semantic reasoning tasks that take sentence pair as input.

In addition, to further incorporating the document-level knowledge, NSP and PSP are extended with non-successive sentences, where the label smoothing technique is adopted.
The proposed method yields a considerable improvement in our experiments.
We evaluate the ability of semantic reasoning on standard NLI and MRC benchmarks, including the challenging HANS dataset~\footnote{Heuristic Analysis for NLI Systems}~\cite{hans}.
Analytical work on the HANS dataset provides a more comprehensible perspective towards the proposed method.
Furthermore, the results on the Chinese benchmarks are provided to demonstrate its generality.

In summary, this work makes the following contributions:
\begin{itemize}
    \item The supervision signal from the original NSP task is weak for semantic inference. Therefore, a novel method is proposed to remedy the asymmetric issue and enhance the reasoning ability.
    \item Both empirical and analytical evaluations are provided on the NLI and MRC datasets, which verifies the effectiveness of using more document-level knowledge.
\end{itemize}

\section{Related Work}

\paragraph{Pair-wise semantic reasoning}
Many NLU tasks seek to model the relationship between two sentences.
Semantic reasoning is performed on the sentence pair for the task-specific inference.
Pair-wise semantic reasoning tasks have drawn a lot of attention from the NLP community as they largely require the comprehension ability of the learning systems.
Recently, the significant improvement on these benchmarks comes from the pre-training models, e.g., BERT, StructBERT~\cite{wang2019structbert}, ERNIE~\cite{ERNIE,ERNIE2}, RoBERTa~\cite{roberta} and XLNet~\cite{xlnet}.
These models learn from unsupervised/self-supervised objectives and perform excellently in the downstream tasks.
Among these models, BERT adopts NSP as one of the objectives in the pre-training and shows that the NSP task has a positive effect on the NLI and MRC tasks.
Although the primary study of XLNet and RoBERTa suggests that NSP is ineffective when the model is trained with a large sequence length of 512, the effect of NSP on the NLI problems should still be emphasized. 
The inefficiency of NSP is likely because the expected context length will be halved for Masked LM when taking a sentence pair as the input.
The models derived from BERT, e.g., StructBERT and ERNIE 1.0/2.0, aim to incorporating more knowledge by elaborating pre-training objectives.
This work aims to enhance the NSP task and verifies whether document-level information is helpful for the pre-training.
To probe whether our method achieves a better regularization ability, our approach is also evaluated on the HANS~\cite{hans} dataset, which contains hard data samples constructed by three heuristics.
Previous advanced models such as BERT fail on the HANS dataset, and the test accuracy can barely exceed 0\% in the subset of test examples.

\paragraph{Unsupervised learning from document}
In recent years, many unsupervised pre-training methods have been proposed in the NLP fields to extract knowledge among sentences~\shortcite{DBLP:conf/nips/KirosZSZUTF15,DBLP:conf/emnlp/ConneauKSBB17,DBLP:conf/iclr/LogeswaranL18,DBLP:journals/corr/abs-1903-09424}.
The prediction of surrounding sentences endows the model with the ability to model the sentence-level coherence.
Skip-Thought~\cite{DBLP:conf/nips/KirosZSZUTF15} consists of an encoder and two decoders. 
When a sentence is given and encoded into a vector by the encoder, the decoders are trained to predict the next sentence and the previous sentence.
The goal is to obtain a better sentence representation that is useful for reconstructing the surrounding context.
Considering that the estimation of the likelihood of sequences is computationally expensive and time-consuming, the Quick-Thought method~\cite{DBLP:conf/iclr/LogeswaranL18} simplifies this in a manner similar to sampled softmax~\cite{DBLP:conf/acl/JeanCMB15}, which classifies the input sentences between surrounding sentences and the other.
Note that Quick-Thought does not distinguish between the previous and next sentence as it is functionally rotation invariant.
However, BERT is order-dependent, and the discrimination can provide more supervision signal for semantic learning.
InferSent~\cite{DBLP:conf/emnlp/ConneauKSBB17} instead pre-trains the model in a manner of supervised learning.
It uses a large-scale NLI dataset as the pre-training task to learn the sentence representation.
In our work, we focus on designing a more effective document-level objective, extended from the NSP task.
The proposed method will be described in the following section and validated by providing extensive experimental results in the experiment part.

\section{Method}

\begin{figure*}[!ht]
    \centering
    \includegraphics[width=6in, height=1.8in]{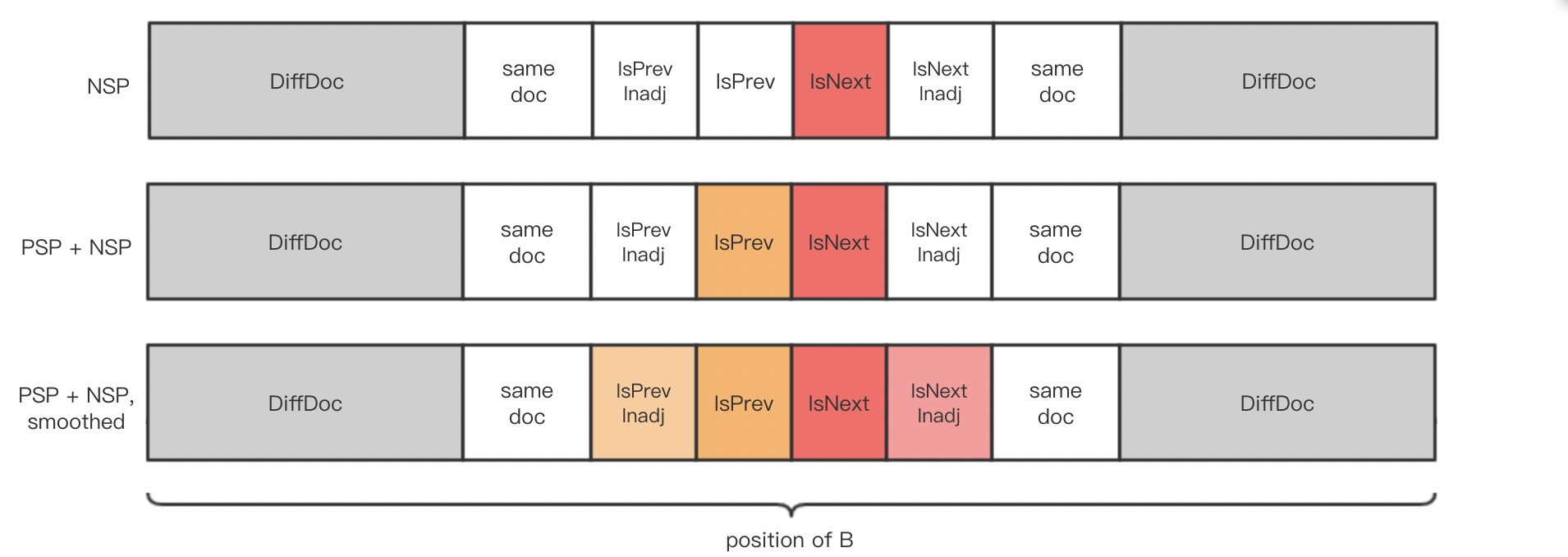}
    \caption{An illustration of the proposed method. \texttt{B} denotes the second input sentence. (1) Top: original NSP task. (2) Middle: 3-class categorization task with \texttt{DiffDoc}, \texttt{IsNext} and \texttt{IsPrev}. (3) Bottom: 3-class task, but with a wider scope of NSP and PSP. The in-adjacent sentences are assisted with a label smoothing technique to reduce the noise.}
    \label{fig:framework}
\end{figure*}

Our method follows the same input format and the model architecture with original BERT.
The proposed method solely concerns the NSP task.
The NSP task is a binary classification task, which takes two sentences (\texttt{A} and \texttt{B}) as input and determines whether \texttt{B} is the next sentence of \texttt{A}.
Although it has been proven to be very effective for BERT, there are two major deficiencies.
(1) Discrimination between \texttt{IsNext} and \texttt{DiffDoc} (the label of the sentences drawn from different documents via negative sampling) is semantically shallow as the signal of sentence order is absent. The correlation between two successive sentences could be obvious, due to, for example, lexical overlap or the conjunction used at the beginning of the second sentence. As reported~\cite{devlin2018bert}, the final pre-trained model is able to achieve 97\%-98\% accuracy on the NSP task.
(2) BERT is order-sensitive, i.e., $f_{\text{BERT}}(
\texttt{A}, \texttt{B}) \neq f_{\text{BERT}}(\texttt{B}, \texttt{A})$, while NSP is uni-directional. 
When the order of the input NLI pair is reversed, the performance will degrade. 
For instance, the accuracy decreases by about 0.5\% on MNLI~\cite{multinli} and 0.4\% on QNLI after swapping the sentences in our experiments~\footnote{The comparison was conducted for 5 times, and the averaged gap is reported.}.

Motivated by these problems, we propose to extend the NSP task with previous sentence prediction (PSP).
Despite its simplicity, empirical results show that this is beneficial for downstream tasks, including both NLI and MRC tasks. 
To further incorporate the document-level information, the scope is also expanded to include more surrounding sentences, not just the adjacent.
The method is briefly illustrated in Fig.~\ref{fig:framework}.

\subsection{Previous Sentence Prediction}
Learning to recognize the previous sentence enables the model to capture more compact context information.
One would argue that \texttt{IsPrev} (the label of PSP) is redundant as it plays a similar role of \texttt{IsNext} (the label of NSP). 
In fact, Quick-Thought uses the sampled softmax to approximate the sentence likelihood estimation of Skip-Thought, and it actually does not differentiate between the previous and next sentences.
However, we suggest the order discrimination is essential for BERT pre-training.
Quick-Thought aims at extracting sentence embedding, and it uses a rotating symmetric function, which makes \texttt{IsPrev} redundant in Quick-Thought.
In contrast, BERT is order-sensitive, and learning the symmetric regularization is rather necessary.
Another advantage of PSP is to enhance document-level supervision.
In order to tell the difference between NSP and PSP, the model has to extract the detailed semantic for inference.

\subsection{Gathering More Document-level Information}
Beyond NSP and PSP, which enable the model to learn the short-term dependency between sentences, we also propose to expand the scope of discrimination task to further incorporate the document-level information.

Specifically, we also include the in-adjacent sentences in the sentence-pair classification task.
The in-adjacent sentences next to the \texttt{IsPrev} and \texttt{IsNext} sentences are sampled, labeled as \texttt{IsPrevInadj} and \texttt{IsNextInadj} (cf. the bottom of Fig.~\ref{fig:framework}).
Note that these in-adjacent sentences will introduce much more training noise to the model.
Therefore, the label smoothing technique is adopted to reduce the noise of these additional samples. 
It achieves this by relaxing our confidence on the labels, e.g., transforming the target probability from (1.0, 0.0) to (0.8, 0.2) in a binary classification problem.

In summary, when \texttt{A} is given, the pre-training example for each label is constructed as follows:
\begin{itemize}
    \item \texttt{IsNext}: Choosing the adjacent following sentence as \texttt{B}.
    \item \texttt{IsPrev}: Choosing the adjacent previous sentence as \texttt{B}.
    \item \texttt{IsNextInadj}: Choosing the in-adjacent following sentence as \texttt{B}. There is a sentence between \texttt{A} and \texttt{B}.
    \item \texttt{IsPrevInadj}: Choosing the in-adjacent previous sentence as \texttt{B}. There is a sentence between \texttt{A} and \texttt{B}.
    \item \texttt{DiffDoc}: Drawing \texttt{B} randomly from a different document.
\end{itemize}

\section{Experiment Settings}
\label{sec:exp}

This section gives detailed experiment settings.
The method is evaluated on the BERTbase model, which has 12 layers, 12 self-attention heads with a hidden size of 768. 

To accelerate the training speed, two-phase training~\cite{devlin2018bert} is adopted.
The first phase uses a maximal sentence length of 128, and 512 for the second phase.
The numbers of training steps of two phases are 50K and 40K for the BERTBase model. 
We used AdamW~\cite{adamw} optimizer with a learning rate of 1e-4, a $\beta_1$ of 0.9, a $\beta_2$ of 0.999 and a L2 weight decay rate of $0.01$.
The first 10\% of the total steps are used for learning rate warming up, followed by the linear decay schema.
We used a dropout probability of 0.1 on all layers. 
The data used for pre-training is the same as BERT, i.e., English Wikipedia (2500M words) and BookCorpus (800M words)~\cite{bookcorpus}. 
For the Masked LM task, we followed the same masking rate and settings as in BERT.

We explore three method settings for comparison.
\begin{itemize}
    \item BERT-PN: The NSP task in BERT is replaced by a 3-class task with \texttt{IsNext}, \texttt{IsPrev} and \texttt{DiffDoc}. The label distribution is 1:1:1.
    \item BERT-PN5cls: The NSP task in BERT is replaced by a 5-class task with two additional labels \texttt{IsNextInadj}, \texttt{IsPrevInadj}. The label distribution is 1:1:1:1:1.
    \item BERT-PNsmth: It uses the same data with BERT-PN5cls, except that the \texttt{IsPrevInadj} (\texttt{IsNextInadj}) label is mapped to \texttt{IsPrev} (\texttt{IsNext}) with a label smoothing factor of 0.8.
\end{itemize}
BERT-PN is used to verify the feasibility of PSP.
The comparison with BERT-PN5cls  illustrates whether more document-level information helps.
BERT-PNsmth, which is the label-smoothed version of BERT-PN5cls, is used to compare with BERT-PN5cls to see whether the noise reduction is necessary.

In the following, we first show that BERT is order-sensitive and the use of PSP remedies this problem.
Then we provide experimental results on the NLI and MRC tasks to verify the effectiveness of the proposed method.
At last, the proposed method is evaluated on several Chinese datasets.

\section{Order-invariant with PSP}
NSP in the pre-training is useful for NLI and MRC task~\cite{devlin2018bert}.
However, we suggested that BERT trained with NSP is order-sensitive, i.e., the performance of BERT depends on the order of the input sentence pair.
To verify our assumption, a primary experiment was conducted.
The order of the input pair of NLI samples is reversed in the fine-tuning phase, and other hyper-parameters and settings keep the same with the BERT paper.
Table~\ref{tab:reverse} shows the accuracy on the validation set of the MNLI~\footnote{The matched set is used for evaluation.} and QNLI datasets.
For the BERTBase model, when the sentences are swapped, the accuracy decreases by 0.5\% on the MNLI task and 0.4\% on the QNLI task.
These results confirm that BERT trained with NSP only is indeed affected by the input order.
This phenomenon motivates us to make the NSP task symmetric.
The results of BERT-PN verify that BERT-PN is order-invariant.
When the input order is reversed, the performance of BERT-PN remains stable.
These results indicate that our method is able to remedy the order-sensitivity problem.

\begin{table}[htp]
    \centering
    \begin{tabular}{c|c|c|c}
    \toprule
    Task & Model     & P\&H & H\&P (reversed) \\
    \hline
    \multirow{2}{*}{MNLI} & BERTBase     & 91.5 & 91.0 (\textbf{-0.5})\\
    & BERTBase-PN  & 91.9 & 92.0 (+0.1)\\
    \hline
    \multirow{2}{*}{QNLI} & BERTBase     & 84.4 & 84.0 (\textbf{-0.4})\\
    & BERTBase-PN  & 85.0 & 84.9 (-0.1)\\
    \toprule
    \end{tabular}
    \caption{The accuracy of BERT and BERT-PN on the validation set of the MNLI and QNLI dataset. P\&H denotes that the input is (premise, hypothesis), which is the order used in BERT. The reported accuracy is the average after 5 runs.}
    \label{tab:reverse}
\end{table}

\section{Results of NLI Tasks}
\subsection{GLUE}
A popular benchmark for evaluation of language understanding is GLUE~\cite{glue}, which is a collection of three NLI tasks (MNLI, QNLI and RTE), three semantic textual similarity (STS) tasks (QQP, STS-B and MRPC), two text classification (TC) tasks (SST-2 and CoLA).
Although the method is motivated for pair-wise reasoning, the results of other problems are also listed.

Our implementation follows the same way that BERT performs in these tasks.
The fine-tuning was conducted for 3 epochs for all the tasks, with a learning rate of 2e-5.
The predictions were obtained by evaluating the training checkpoint with the best validation performance.

Table~\ref{tab:glue} illustrates the experimental results, showing that our method is beneficial for all of NLI tasks.
The improvement on the RTE dataset is significant, i.e., 4\% absolute gain over the BERTBase.
Besides NLI, our model also performs better than BERTBase in the STS task.
The STS tasks are semantically similar to the NLI tasks, and hence able to take advantage of PSP as well.
Actually, the proposed method has a positive effect whenever the input is a sentence pair.
The improvements suggest that the PSP task encourages the model to learn more detailed semantics in the pre-training, which improves the model on the downstream learning tasks.
Moreover, our method is surprisingly able to achieve slightly better results in the single-sentence problem.
The improvement should be attributed to better semantic representation.

When comparing between PN and PN5cls, PN5cls achieves better results than PN.
This indicates that including a broader range of the context is effective for improving  inference ability.
Considering that the representation of \texttt{IsNext} and \texttt{IsNextInadj} should be coherent, 
we propose BERTBase-PNsmth to mitigate this problem.
PNsmth further improves the performance and obtains an averaged score of 81.0.

\begin{table*}[htp]
\small
    \centering
    \begin{tabular}{c|c|c|c|c|c|c|c|c|c}
    \toprule
                 & \multicolumn{3}{c|}{NLI} & \multicolumn{3}{c|}{STS} & \multicolumn{2}{c|}{TC} & \\
                 &MNLI          & QNLI  & RTE   & QQP   & STS-B & MRPC  & SST-2 & CoLA & Average\\
                 &392k          & 108k  & 2.5k  & 363k  & 8.5k  & 3.5k  & 67k   & 5.7k & -      \\ \midrule
BiLSTM+ELMo+Attn &76.4/76.1     & 79.8  & 64.8  & 56.8  & 73.3  & 84.9  & 90.4  & 36.0 & 71.0   \\
OpenAI GPT       &82.1/81.4     & 87.4  & 56.0  & 70.3  & 80.0  & 82.3  & 91.3  & 45.4 & 75.1   \\
BERTBase         &84.6/83.4     & 90.5  & 66.4  & 71.2  & 85.8  & 88.9  & 93.5  & 52.1 & 79.6   \\ \midrule 
BERTBase-PN      &84.2/84.1     & 92.2  & 70.2  & 71.7  & 87.2  & 88.9  & \textbf{94.2}  & 51.1 & 80.4\\
BERTBase-PN5cls  &84.6/84.3            & \textbf{92.3} & 70.0          & 71.9          & \textbf{87.5}  & \textbf{89.8}          & 93.5          & 52.0          & 80.7 \\
BERTBase-PNsmth  &\textbf{85.2}/\textbf{84.4}   & 92.1          & \textbf{70.6} & \textbf{72.2} & 86.4  & \textbf{89.8} & \textbf{94.2} & \textbf{54.6} &  \textbf{81.0}  \\

    \toprule
    \end{tabular}
    \caption{Results on the test set of GLUE benchmark. The performance was obtained by the official evaluation server. The number below each task is the number of training examples. 
    The "Average" column follows the setting in the BERT paper, which excludes the problematic WNLI task.
    F1 scores are reported for QQP and MRPC, Spearman correlations are reported for STS-B, and accuracy scores are reported for the other tasks.
    All the listed models are trained on the Wikipedia and the Book Corpus datasets. The results are the average of 5 runs.
    }
    \label{tab:glue}
\end{table*}

\subsection{HANS}
Although BERT has shown its effectiveness in the NLI tasks.
\citeauthor{hans} pointed out that BERT is still vulnerable in the NLI task as it is prone to adopting fallible heuristics. 
Therefore, they released a dataset, called The Heuristic Analysis for NLI Systems (HANS), to probe whether the model learns inappropriate inductive bias from the training set.
It is constructed by three heuristics, 
i.e., lexical overlap heuristic, sub-sequence heuristic, and constituent heuristic. 
The first heuristic assumes that a premise entails all hypotheses constructed from words in the premise, the second assumes that a premise entails all of its contiguous sub-sequences and the third assumes that a premise entails all complete sub-trees in its parse tree.
BERT and other advanced models fail on this dataset and barely exceeds 0\% accuracy in most cases~\cite{hans}.

\begin{figure}[!htb]
    \centering
    \includegraphics[width=0.5\textwidth,height=0.30\textheight]{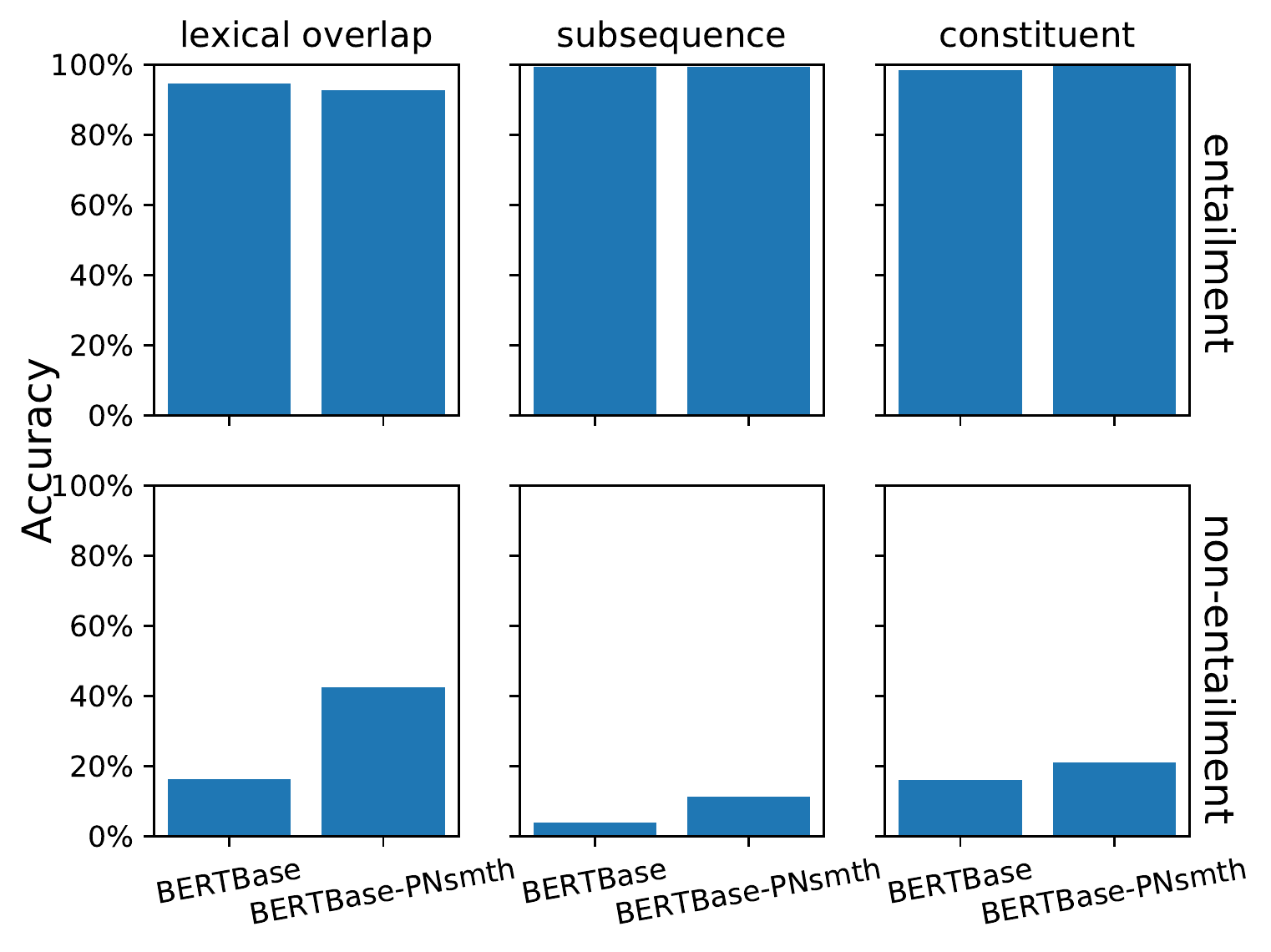}
    \caption{The accuracy on evaluation set of HANS. It has six sub-components, each defined by its correct label and the heuristic it addresses. 
    }
    \label{fig:hans}
\end{figure}

\begin{figure*}
    \centering
    \includegraphics[width=0.8\textwidth,height=0.38\textheight]{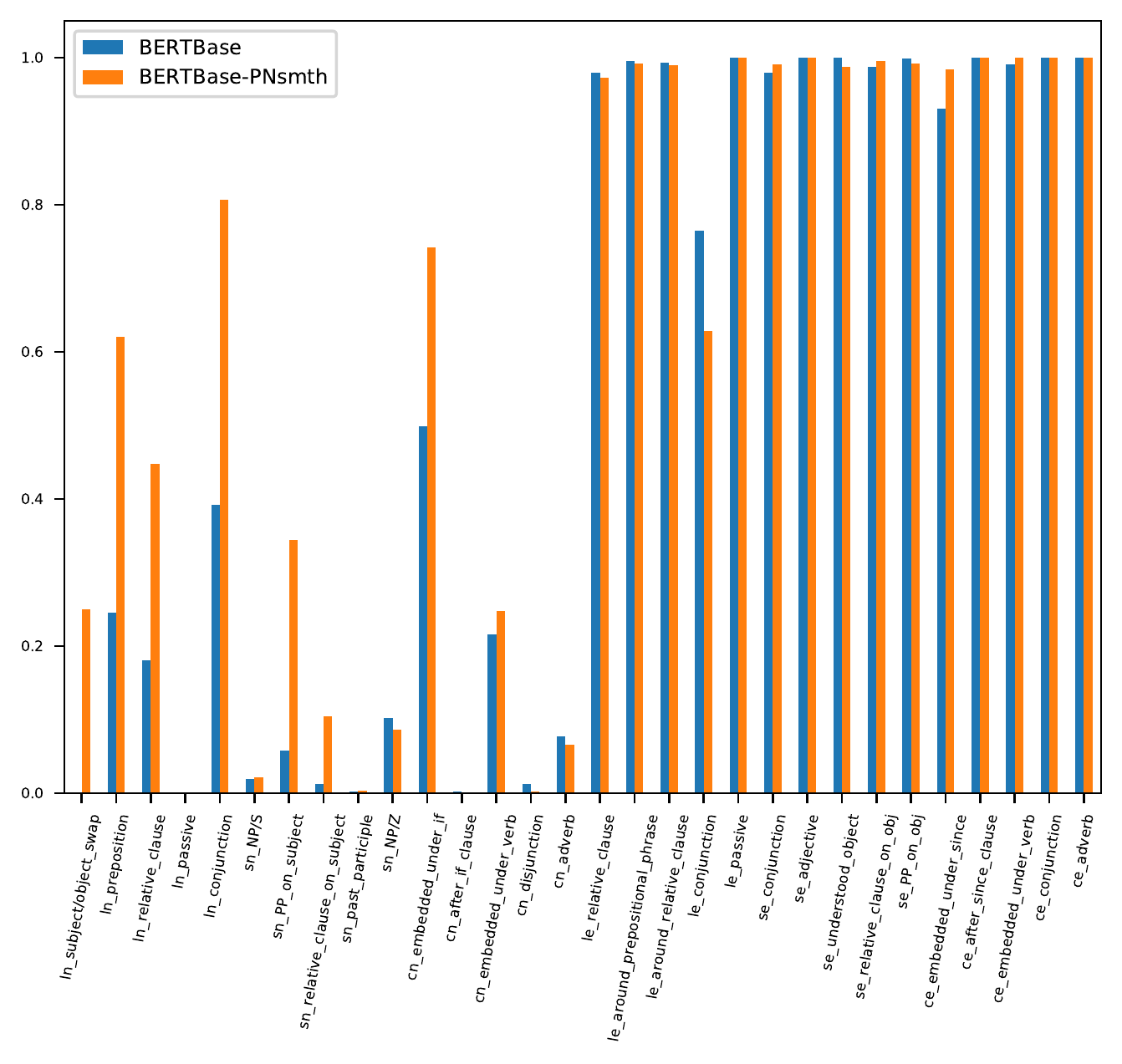}
    \caption{Performance on thirty detailed sub-components of the HANS evaluation set (30K instances). 
    Each sub-component is defined by three heuristics, i.e., Lexical overlap, Sub-sequence and Constituent. For instance, in prefix ``ln'' , ``l'' denotes lexical overlap heuristic, ``n'' denotes the \texttt{non-entailment} label. 
    The suffix means a specific syntactic rule, e.g., subject/object\underline{ }swap means in the hypothesis sentence, the subject and the object are swapped.}
    \label{fig:hans_details}
\end{figure*}

Fig.~\ref{fig:hans} illustrates the accuracy of BERTBase and BERTBase-PNsmth on the HANS dataset.
The evaluation is made upon the model trained on the MNLI dataset and the predicted \texttt{neutral} and \texttt{contradiction} labels are mapped into \texttt{non-entailment}.
The BERTBase-PNsmth evidently outperforms the BERTBase with the \texttt{non-entailment} examples.
For the \texttt{non-entailment} samples constructed using the lexical overlap heuristic, our model achieves 160\% relative improvement over the BERTBase model.
Some samples are constructed by swapping the entities in the sentence (e.g., \emph{The doctor saw the lawyer} $\nrightarrow$ \emph{The lawyer saw the doctor}) and our method outperforms BERTBase by 20\% in accuracy.
We suggest that the Masked LM task can hardly model the relationship between two entities and NSP only is too semantically shallow to capture the precise meaning.
However, the discrimination between NSP and PSP enhances the model to realize the role of entities in a given sentence.
For example, to determine that \texttt{A} (\emph{X is beautiful}) rather than $\bar{\texttt{A}}$ (\emph{Y is beautiful}) is the previous sentence of \texttt{B} (\emph{Y loves X}),  the model have to recognize the relationship between \emph{X} and \emph{Y}. 
In contrast, when PSP is absent, NSP can be probably inferred by learning the occurrence between \emph{beautiful} and \emph{loves}, regardless of the sentence structure.
The detailed performance of the proposed method on the HANS dataset is illustrated in Fig.~\ref{fig:hans_details}.
The definition of each heuristic rules can be found in ~\cite{hans}.

\section{Results of MRC Tasks}
\subsection{SQuAD v1.1 and v2.0}
We also evaluate our method on the MRC tasks.
The Stanford Question Answering Dataset (SQuAD v1.1) is a question answering (QA) dataset, which consists of 100K samples~\cite{DBLP:conf/emnlp/RajpurkarZLL16}.
Each data sample has a question and a corresponding Wikipedia passage that contains the answer.
The goal is to extract the answer from the passage for the given question.

In the fine-tuning procedure, we follow the exact way the BERT performed.
The output vectors are used to compute the score of tokens being start and end of the answer span.
The valid span that has the maximum score is selected as the prediction.
And similarly, the fine-tuning training was performed for 3 epochs with a learning rate of 3e-5.

Table~\ref{tab:squad1.1} demonstrates the results on the SQuAD v1.1 dataset. 
The comparison between BERTBase-PN and BERTBase indicates that the inclusion of the PSP subtask is beneficial (2.4\% absolute improvement).
When using BERTBase-PNsmth, another 0.3\% increase in EM can be obtained.
The experimental results on the SQuAD v2.0~\cite{DBLP:conf/acl/RajpurkarJL18} are also shown in Table.~\ref{tab:squad1.1}.
The SQuAD v2.0 differs from SQuAD v1.1 by allowing the question-paragraph pairs that have no answer.
For SQuAD v2.0, our method also achieved about 4\% absolute improvement in both EM and F1 against BERTBase.

\begin{table}
\small
    \centering
    \begin{tabular}{c|cc|cc}
    \toprule
    \multirow{2}{*}{Model} & \multicolumn{2}{c|}{Dev v1.1} & \multicolumn{2}{c}{Dev v2.0} \\
    & EM & F1 & EM & F1 \\
    \hline
       RoBERTaBase & - & \textbf{90.6} & - & 79.7 \\
       BERTBase             & 80.8 & 88.5 & 72.8 & 76.3\\
       BERTBase-PN          & 83.2 & 90.5 & 76.5 & 79.6\\
       BERTBase-PN5cls      & 83.3 & \textbf{90.6} & 77.0 & 80.3\\
       BERTBase-PNsmth      & \textbf{83.6} & \textbf{90.6}& \textbf{77.4} & \textbf{80.6} \\
    \toprule
    \end{tabular}
    \caption{The performance of various BERT models fine-tuned on the SQuAD v1.1 and v2.0 dataset. EM means the percentage of exact match.
    The results of RoBERTa is the DOC-SENTENCES version retrieved from Table 2 in ~\cite{roberta}.
    }
    \label{tab:squad1.1}
\end{table}

\subsection{RACE}
The ReAding Comprehension from Examinations (RACE) dataset~\cite{DBLP:conf/emnlp/LaiXLYH17} consists of 100K questions taken from English exams, and the answers are generated by human experts.
This is one of the most challenging MRC datasets that require sophisticated reasoning.

In our implementation, the question, document, and option are concatenated as a single sequence, separated by \texttt{[SEP]} token.
And each part is truncated by a maximal length of 40/432/40, respectively.
The model computes for a concatenation a scalar as the score, which is then used in a softmax layer for the final prediction.
The fine-tuning was conducted for 5 epochs, with a batch size of 32 and a learning rate of 5e-5.
As shown in Table~\ref{tab:race}, the proposed method significantly improve the performance on the RACE dataset.
BERTBase-PN obtains 2.6\% accuracy improvement, and BERTBase-PN5cls further brings 0.4\% absolute gain.

The comparisons on the SQuAD v1.1, SQuAD v2.0, and RACE dataset demonstrate that the involvement of additional sentence and discourse information is not only beneficial for the NLI task but also the MRC task.
This is reasonable as these tasks heavily rely on the global semantic understanding and  sophisticated reasoning among sentences.
And this ability can be effectively enhanced by our method.

\begin{table}
\small
    \centering
    \begin{tabular}{c|c|c|c}
    \toprule
      Model             & Middle & High & Accuracy \\
    \hline
      RoBERTaBase  & - & - & 65.6 \\
      BERTBase          & 71.8   & 63.6 & 66.0     \\
      BERTBase-PN       & 74.2   & 66.3 & 68.6     \\
      BERTBase-PN5cls   & \textbf{75.8}   & 66.2 & \textbf{69.0}     \\
      BERTBase-PNsmth   & 74.1   & \textbf{66.3} & 68.6          \\
    \toprule
    \end{tabular}
    \caption{The experimental results on test set of the RACE dataset. 
    The results of RoBERTa is the DOC-SENTENCES version retrieved from Table 2 in ~\cite{roberta}.
    All the listed models are trained on the Wikipedia and the Book Corpus datasets.}
    \label{tab:race}
\end{table}

\section{Results of Chinese NLP Tasks}

\begin{table*}[!htb]
\small
\centering
\begin{tabular}{@{}c|l|r|r|r|r|r|r@{}}
\toprule
             & \multirow{2}{*}{metrics}  & BERT    & BERT-wwm & ERNIE  & ERNIE2.0 & BERT-PN & BERT-PNsmth \\ 
             &      & 393M    & 393M     & -      & 14988M   & 10879M  & 10879M \\ \midrule
\multicolumn{8}{l}{\textit{Single-task single base models on dev}}                                                  \\
XNLI         & Accuracy & 77.8 (77.4) & 79.0 (78.4)  & - (79.9) & - (\textbf{81.2})       & 80.5 (79.9) & \textbf{81.4} (81.0)     \\
LCQMC        & Accuracy & 89.4 (88.4) & 89.4 (89.2)  & - (89.7) & - (\textbf{90.9})       & 90.3 (89.4) & \textbf{90.6} (90.1)     \\
NLPCC-DBQA   & F1       & - (80.7)     & - (-)        & - (82.3)    & - (84.7)       & 85.0 (84.6) & \textbf{85.9} (\textbf{85.4})     \\ \midrule
\multicolumn{8}{l}{\textit{Single-task single base models on test}}                                                 \\
XNLI         & Accuracy & 77.8 (77.5) & 78.2 (78.0)  & - (78.4) & - (79.7)       & 79.8 (79.4) & \textbf{80.3} (\textbf{79.9})     \\
LCQMC        & Accuracy & 86.9 (86.4) & 87.0 (86.8)  & - (87.4) & - (87.9)       & \textbf{88.7} (87.5) & \textbf{88.7} (\textbf{88.0})     \\
NLPCC-DBQA   & F1       & - (80.8)    & - (-)        &- (82.7)    & - (85.3)       & 85.2 (84.9) & \textbf{86.2} (\textbf{85.9})     \\ \bottomrule
\end{tabular}
\caption{
Comparison on the Chinese NLP tasks. 
All the models are of ``base'' size. 
The results of BERT, BERT-wwm are retrieved from literature~\cite{wwm}, except the results of NLPCC-DBQA which is from ERNIE 2.0~\shortcite{ERNIE2}. 
The results of ERNIE, ERNIE 2.0 are retrieved from literature~\cite{ERNIE,ERNIE2}.
The best result and the average (in bracket) of 5 runs are reported.
The number below the model denotes the number of tokens in the pre-training data.
}
\label{tab:chinese}
\end{table*}

\begin{table*}[!htb]
\small
\centering
\begin{tabular}{@{}l|rr|rr|rr@{}}
\toprule
& \multicolumn{2}{c|}{CMRC-2018 (Dev)}   & \multicolumn{2}{c|}{DRCD (Dev)}  & \multicolumn{2}{c}{DRCD (Test)}  \\ \midrule
metrics          & F1          & EM          & F1          & EM          & F1          & EM          \\ \midrule
BERTBase (ours)  & 84.7 (84.3)  & 64.1 (63.8)  & 90.2 (90.0)  & 83.5 (83.4)  & 89.0 (88.9)  & 82.0 (81.8)  \\
BERTBase~\cite{wwm}   & 84.5 (84.0) & 65.5 (64.4) & 89.9 (89.6) & 83.1 (82.7) & 89.2 (88.8) & 82.2 (81.6) \\
BERTBase~\cite{ERNIE2} & - (85.9)       & - (66.3)       & - (91.6)       & - (85.7)       & - (90.9)       & - (84.9)       \\
BERTBase-wwm     & 85.6 (84.7) & 66.3 (65.0) & 90.5 (90.2)  & 83.7 (83.5)  & 89.8 (89.4)  & 82.7 (82.1)  \\
BERTBase-PN  & \textbf{87.5} (\textbf{86.8})  & \textbf{66.6} (65.8)  &   92.3 (92.0) &   86.4 (86.0)              &     92.3 (92.2)       &   86.1 (86.0)          \\
BERTBase-PNsmth      & 86.4 (86.2)  & 66.5 (\textbf{66.3})  & \textbf{93.0} (\textbf{92.7})  & \textbf{86.8} (\textbf{86.8})  & \textbf{92.6} (\textbf{92.5})  & \textbf{86.7} (\textbf{86.6})  \\
\bottomrule
\end{tabular}
\caption{Results on the CMRC-2018 and DRCD datasets.
Three BERTBase models are reported from our reproduction, BERT-wwm paper~\cite{wwm} and ERNIE 2.0 paper~\cite{ERNIE2}, respectively.
The results of BERTBase-wwm are obtained from the paper~\cite{wwm}.
EM denotes the percentage of exact matching. 
The best result and the average (in bracket) of 5 runs are reported.
}
    \label{tab:chineseqa}
\end{table*}

The experiments are also conducted on Chinese NLP tasks:
\begin{itemize}
    \item XNLI~\cite{xnli} a multi-lingual dataset. The data sample in XNLI is a sentence pair annotated with textual entailment. The Chinese part is used.
    \item LCQMC~\cite{lcqmc} is a dataset for sequence matching. A binary label is annotated for a sentence pair in the dataset to indicate whether these two sentences have the same intention. 
    \item NLPCC-DBQA~\cite{nlpccdbqa} formulates the domain-based question answering as a binary classification task. Each data sample is a question-sentence pair. The goal is to identify whether the sentence contains the answer to the question.
    \item CMRC-2018~\footnote{https://hfl-rc.github.io/cmrc2018} is the Chinese Machine Reading Comprehension dataset. Similar to SQuAD, the system needs to extract fragments from the text as the answer.
    \item DRCD~\cite{drcd} is also a Chinese MRC data set. The data follows the format of SQuAD.
\end{itemize}

For Chinese NLP tasks, we pre-train the model using Chinese corpus.
We collected textual data (10879M tokens in total) from the website, consisting of  Hudong Baike data (6084M tokens)~\footnote{http://www.baike.com}, Zhihu data(465M tokens)~\footnote{http://www.zhihu.com}, Sohu News(3937M tokens)~\footnote{http://news.sohu.com} and Wikipedia data (393M tokens).

For the first 3 Chinese tasks, we follow the settings  as in ERNIE~\cite{ERNIE}.
The experimental results are given in Table~\ref{tab:chinese}.
The proposed method is compared with four models, i.e., BERTBase~\cite{devlin2018bert}, BERTBase with whole word masking~\cite{wwm}, ERNIE~\cite{ERNIE} and ERNIE 2.0~\cite{ERNIE2}.
Our method achieves comparable or even better results against ERNIE 2.0~\cite{ERNIE2}.
Note that the Chinese ERNIE 2.0 is equipped with 5 different objectives and it uses more training data (14988M tokens in total) than ours.
The results indicate that the proposed method is quite effective for the pair-wise semantic reasoning as simply including PSP can achieve the results on par with multiple objectives.

The results of CMRC-2018 and DRCD datasets are given in Table~\ref{tab:chineseqa}.
Since the CMRC-2018 competition does not release the test set, the comparison on the test set is absent.
Our results are obtained using the open-sourced code of BERT-wwm~\footnote{ https://github.com/ymcui/cmrc2018/tree/master/baseline}.
We keep the hyper-parameters the same with that in ERNIE~\cite{ERNIE}, except that the batch size is 12 instead of 64 due to the memory limit.
Under this setting, we achieved similar results of BERTBase in the BERT-wwm paper~\cite{wwm}.
However, this is worse than the results of BERTBase reported in the ERNIE 2.0 paper~\cite{ERNIE2} by about 1\% in F1.
This suggests that our results are currently incomparable with ERNIE 2.0.
Overall, the results in Table~\ref{tab:chineseqa} illustrate that our method is also effective for the Chinese QA tasks.

\section{Conclusion}
This paper aims to enrich the NSP task to provide more document-level information in the pre-training.
Motivated by the in-symmetric property of NSP, we propose to differentiate between different sentence orders by including PSP.
Despite the simplicity, extensive experiments demonstrate that the model obtains a better ability in pair-wise semantic reasoning.
Our work suggests that the document-level objective is effective, at least for the BERTbase model.
In the future, we will investigate the way to take advantages of both large-scale training and our method.



\bibliographystyle{aaai}
\bibliography{references}

\begin{thebibliography}{}

\bibitem[\protect\citeauthoryear{Conneau \bgroup et al\mbox.\egroup
  }{2017}]{DBLP:conf/emnlp/ConneauKSBB17}
Conneau, A.; Kiela, D.; Schwenk, H.; Barrault, L.; and Bordes, A.
\newblock 2017.
\newblock Supervised learning of universal sentence representations from
  natural language inference data.
\newblock In {\em {EMNLP}},  670--680.
\newblock Association for Computational Linguistics.

\bibitem[\protect\citeauthoryear{Conneau \bgroup et al\mbox.\egroup
  }{2018}]{xnli}
Conneau, A.; Rinott, R.; Lample, G.; Williams, A.; Bowman, S.~R.; Schwenk, H.;
  and Stoyanov, V.
\newblock 2018.
\newblock Xnli: Evaluating cross-lingual sentence representations.
\newblock In {\em Proceedings of the 2018 Conference on Empirical Methods in
  Natural Language Processing}.
\newblock Association for Computational Linguistics.

\bibitem[\protect\citeauthoryear{Cui \bgroup et al\mbox.\egroup }{2019}]{wwm}
Cui, Y.; Che, W.; Liu, T.; Qin, B.; Yang, Z.; Wang, S.; and Hu, G.
\newblock 2019.
\newblock Pre-training with whole word masking for chinese {BERT}.
\newblock {\em CoRR} abs/1906.08101.

\bibitem[\protect\citeauthoryear{Devlin \bgroup et al\mbox.\egroup
  }{2018}]{devlin2018bert}
Devlin, J.; Chang, M.-W.; Lee, K.; and Toutanova, K.
\newblock 2018.
\newblock {BERT:} pre-training of deep bidirectional transformers for language
  understanding.
\newblock {\em arXiv preprint arXiv:1810.04805}.

\bibitem[\protect\citeauthoryear{Duan and Tang}{2017}]{nlpccdbqa}
Duan, N., and Tang, D.
\newblock 2017.
\newblock Overview of the {NLPCC} 2017 shared task: Open domain chinese
  question answering.
\newblock In {\em Natural Language Processing and Chinese Computing - 6th {CCF}
  International Conference, {NLPCC} 2017, Dalian, China, November 8-12, 2017,
  Proceedings},  954--961.

\bibitem[\protect\citeauthoryear{Jean \bgroup et al\mbox.\egroup
  }{2015}]{DBLP:conf/acl/JeanCMB15}
Jean, S.; Cho, K.; Memisevic, R.; and Bengio, Y.
\newblock 2015.
\newblock On using very large target vocabulary for neural machine translation.
\newblock In {\em Proceedings of the 53rd Annual Meeting of the Association for
  Computational Linguistics and the 7th International Joint Conference on
  Natural Language Processing of the Asian Federation of Natural Language
  Processing, {ACL} 2015, July 26-31, 2015, Beijing, China, Volume 1: Long
  Papers},  1--10.

\bibitem[\protect\citeauthoryear{Kiros \bgroup et al\mbox.\egroup
  }{2015}]{DBLP:conf/nips/KirosZSZUTF15}
Kiros, R.; Zhu, Y.; Salakhutdinov, R.; Zemel, R.~S.; Urtasun, R.; Torralba, A.;
  and Fidler, S.
\newblock 2015.
\newblock Skip-thought vectors.
\newblock In {\em {NIPS}},  3294--3302.

\bibitem[\protect\citeauthoryear{Lai \bgroup et al\mbox.\egroup
  }{2017}]{DBLP:conf/emnlp/LaiXLYH17}
Lai, G.; Xie, Q.; Liu, H.; Yang, Y.; and Hovy, E.~H.
\newblock 2017.
\newblock {RACE:} large-scale reading comprehension dataset from examinations.
\newblock In {\em {EMNLP}},  785--794.
\newblock Association for Computational Linguistics.

\bibitem[\protect\citeauthoryear{Liu \bgroup et al\mbox.\egroup }{2018}]{lcqmc}
Liu, X.; Chen, Q.; Deng, C.; Zeng, H.; Chen, J.; Li, D.; and Tang, B.
\newblock 2018.
\newblock {LCQMC:} {A} large-scale chinese question matching corpus.
\newblock In {\em Proceedings of the 27th International Conference on
  Computational Linguistics, {COLING} 2018, Santa Fe, New Mexico, USA, August
  20-26, 2018},  1952--1962.

\bibitem[\protect\citeauthoryear{Liu \bgroup et al\mbox.\egroup
  }{2019}]{roberta}
Liu, Y.; Ott, M.; Goyal, N.; Du, J.; Joshi, M.; Chen, D.; Levy, O.; Lewis, M.;
  Zettlemoyer, L.; and Stoyanov, V.
\newblock 2019.
\newblock Roberta: {A} robustly optimized {BERT} pretraining approach.
\newblock {\em CoRR} abs/1907.11692.

\bibitem[\protect\citeauthoryear{Logeswaran and
  Lee}{2018}]{DBLP:conf/iclr/LogeswaranL18}
Logeswaran, L., and Lee, H.
\newblock 2018.
\newblock An efficient framework for learning sentence representations.
\newblock In {\em {ICLR}}.
\newblock OpenReview.net.

\bibitem[\protect\citeauthoryear{Loshchilov and Hutter}{2019}]{adamw}
Loshchilov, I., and Hutter, F.
\newblock 2019.
\newblock Decoupled weight decay regularization.
\newblock In {\em 7th International Conference on Learning Representations,
  {ICLR} 2019, New Orleans, LA, USA, May 6-9, 2019}.

\bibitem[\protect\citeauthoryear{McCoy, Pavlick, and Linzen}{2019}]{hans}
McCoy, T.; Pavlick, E.; and Linzen, T.
\newblock 2019.
\newblock Right for the wrong reasons: Diagnosing syntactic heuristics in
  natural language inference.
\newblock In {\em Proceedings of the 57th Conference of the Association for
  Computational Linguistics, {ACL} 2019, Florence, Italy, July 28- August 2,
  2019, Volume 1: Long Papers},  3428--3448.

\bibitem[\protect\citeauthoryear{Rajpurkar \bgroup et al\mbox.\egroup
  }{2016}]{DBLP:conf/emnlp/RajpurkarZLL16}
Rajpurkar, P.; Zhang, J.; Lopyrev, K.; and Liang, P.
\newblock 2016.
\newblock Squad: 100, 000+ questions for machine comprehension of text.
\newblock In {\em Proceedings of the 2016 Conference on Empirical Methods in
  Natural Language Processing, {EMNLP} 2016, Austin, Texas, USA, November 1-4,
  2016},  2383--2392.

\bibitem[\protect\citeauthoryear{Rajpurkar, Jia, and
  Liang}{2018}]{DBLP:conf/acl/RajpurkarJL18}
Rajpurkar, P.; Jia, R.; and Liang, P.
\newblock 2018.
\newblock Know what you don't know: Unanswerable questions for squad.
\newblock In {\em Proceedings of the 56th Annual Meeting of the Association for
  Computational Linguistics, {ACL} 2018, Melbourne, Australia, July 15-20,
  2018, Volume 2: Short Papers},  784--789.

\bibitem[\protect\citeauthoryear{Shao \bgroup et al\mbox.\egroup }{2018}]{drcd}
Shao, C.; Liu, T.; Lai, Y.; Tseng, Y.; and Tsai, S.
\newblock 2018.
\newblock {DRCD:} a chinese machine reading comprehension dataset.
\newblock {\em CoRR} abs/1806.00920.

\bibitem[\protect\citeauthoryear{Sun \bgroup et al\mbox.\egroup
  }{2019a}]{ERNIE}
Sun, Y.; Wang, S.; Li, Y.; Feng, S.; Chen, X.; Zhang, H.; Tian, X.; Zhu, D.;
  Tian, H.; and Wu, H.
\newblock 2019a.
\newblock {ERNIE:} enhanced representation through knowledge integration.
\newblock {\em CoRR} abs/1904.09223.

\bibitem[\protect\citeauthoryear{Sun \bgroup et al\mbox.\egroup
  }{2019b}]{ERNIE2}
Sun, Y.; Wang, S.; Li, Y.; Feng, S.; Tian, H.; Wu, H.; and Wang, H.
\newblock 2019b.
\newblock {ERNIE} 2.0: {A} continual pre-training framework for language
  understanding.
\newblock {\em CoRR} abs/1907.12412.

\bibitem[\protect\citeauthoryear{Wang \bgroup et al\mbox.\egroup
  }{2019a}]{glue}
Wang, A.; Singh, A.; Michael, J.; Hill, F.; Levy, O.; and Bowman, S.~R.
\newblock 2019a.
\newblock {GLUE:} {A} multi-task benchmark and analysis platform for natural
  language understanding.
\newblock In {\em 7th International Conference on Learning Representations,
  {ICLR} 2019, New Orleans, LA, USA, May 6-9, 2019}.

\bibitem[\protect\citeauthoryear{Wang \bgroup et al\mbox.\egroup
  }{2019b}]{wang2019structbert}
Wang, W.; Bi, B.; Yan, M.; Wu, C.; Bao, Z.; Peng, L.; and Si, L.
\newblock 2019b.
\newblock Structbert: Incorporating language structures into pre-training for
  deep language understanding.
\newblock {\em arXiv preprint arXiv:1908.04577}.

\bibitem[\protect\citeauthoryear{Williams, Nangia, and Bowman}{2018}]{multinli}
Williams, A.; Nangia, N.; and Bowman, S.
\newblock 2018.
\newblock A broad-coverage challenge corpus for sentence understanding through
  inference.
\newblock In {\em Proceedings of the 2018 Conference of the North American
  Chapter of the Association for Computational Linguistics: Human Language
  Technologies, Volume 1 (Long Papers)},  1112--1122.
\newblock Association for Computational Linguistics.

\bibitem[\protect\citeauthoryear{Yang \bgroup et al\mbox.\egroup
  }{2019}]{xlnet}
Yang, Z.; Dai, Z.; Yang, Y.; Carbonell, J.~G.; Salakhutdinov, R.; and Le, Q.~V.
\newblock 2019.
\newblock Xlnet: Generalized autoregressive pretraining for language
  understanding.
\newblock {\em CoRR} abs/1906.08237.

\bibitem[\protect\citeauthoryear{Zhou, Cheng, and
  Zhang}{2019}]{DBLP:journals/corr/abs-1903-09424}
Zhou, J.; Cheng, X.; and Zhang, J.
\newblock 2019.
\newblock An end-to-end neural network framework for text clustering.
\newblock {\em CoRR} abs/1903.09424.

\bibitem[\protect\citeauthoryear{Zhu \bgroup et al\mbox.\egroup
  }{2015}]{bookcorpus}
Zhu, Y.; Kiros, R.; Zemel, R.~S.; Salakhutdinov, R.; Urtasun, R.; Torralba, A.;
  and Fidler, S.
\newblock 2015.
\newblock Aligning books and movies: Towards story-like visual explanations by
  watching movies and reading books.
\newblock In {\em 2015 {IEEE} International Conference on Computer Vision,
  {ICCV} 2015, Santiago, Chile, December 7-13, 2015},  19--27.

\end{thebibliography}

\end{document}